\documentclass[sigconf]{acmart}
\AtBeginDocument{%
  \providecommand\BibTeX{{%
    \normalfont B\kern-0.5em{\scshape i\kern-0.25em b}\kern-0.8em\TeX}}}

\copyrightyear{2024}
\acmYear{2024}
\setcopyright{acmlicensed}
\acmBooktitle{Proceedings of}
\acmDOI{10.1145/3664647.3681391}
\acmISBN{979-8-4007-0686-8/24/10}

\newcommand{\et}[2]{${#1}^{\pm{#2}}$}
\newcommand{\etb}[2]{$\mathbf{{#1}}^{\pm{#2}}$}

\begin{document}

%%
%% The "title" command has an optional parameter,
%% allowing the author to define a "short title" to be used in page headers.
\title{DreamVTON: Customizing 3D Virtual Try-on with Personalized Diffusion Models}

%%
%% The "author" command and its associated commands are used to define
%% the authors and their affiliations.
%% Of note is the shared affiliation of the first two authors, and the
%% "authornote" and "authornotemark" commands
%% used to denote shared contribution to the research.

\author{Zhenyu Xie}
\affiliation{%
  \institution{Shenzhen Campus of Sun Yat-sen University}
  \institution{ByteDance}
  % \streetaddress{1 Th{\o}rv{\"a}ld Circle}
  \city{ShenZhen, Guangdong}
  \country{China}}
\email{xiezhy6@mail2.sysu.edu.cn}

\author{Haoye Dong}
\affiliation{%
  \institution{Carnegie Mellon University}
  % \streetaddress{1 Th{\o}rv{\"a}ld Circle}
  \city{Pittsburgh, Pennsylvania}
  \country{United States}}
\email{donghy7@mail2.sysu.edu.cn}

\author{Yufei Gao}
\affiliation{%
  \institution{Shenzhen Campus of Sun Yat-sen University}
  % \streetaddress{1 Th{\o}rv{\"a}ld Circle}
  \city{ShenZhen, Guangdong}
  \country{China}}
\email{gaoyf37@mail2.sysu.edu.cn}

\author{Zehua Ma}
\affiliation{%
  \institution{Shenzhen Campus of Sun Yat-sen University}
  % \streetaddress{1 Th{\o}rv{\"a}ld Circle}
  \city{ShenZhen, Guangdong}
  \country{China}}
\email{mazh58@mail2.sysu.edu.cn}

\author{Xiaodan Liang*}
\affiliation{%
  \institution{Shenzhen Campus of Sun Yat-sen University}
  \institution{Research Institute of Multiple Agents and Embodied Intelligence, Peng Cheng Laboratory}
  % \streetaddress{1 Th{\o}rv{\"a}ld Circle}
  \city{ShenZhen, Guangdong}
  \country{China}}
\email{xdliang328@gmail.com}

%%
%% By default, the full list of authors will be used in the page
%% headers. Often, this list is too long, and will overlap
%% other information printed in the page headers. This command allows
%% the author to define a more concise list
%% of authors' names for this purpose.
\renewcommand{\shortauthors}{Zhenyu Xie, et al.}

%%
%% The abstract is a short summary of the work to be presented in the
%% article.
\begin{abstract}
Image-based 3D Virtual Try-ON (VTON) aims to sculpt the 3D human according to person and clothes images, which is data-efficient (i.e., getting rid of expensive 3D data) but challenging.
Recent text-to-3D methods achieve remarkable improvement in high-fidelity 3D human generation, demonstrating its potential for 3D virtual try-on.
Inspired by the impressive success of personalized diffusion models (e.g., Dreambooth and LoRA) for 2D VTON, it is straightforward to achieve 3D VTON by integrating the personalization technique into the diffusion-based text-to-3D framework.
However, employing the personalized module in a pre-trained diffusion model (e.g., StableDiffusion (SD)) would degrade the model's capability for multi-view or multi-domain synthesis, which is detrimental to the geometry and texture optimization guided by Score Distillation Sampling (SDS) loss.
In this work, we propose a novel customizing 3D human try-on model, named \textbf{DreamVTON}, to separately optimize the geometry and texture of the 3D human.
Specifically, a personalized SD with multi-concept LoRA is proposed to provide the generative prior about the specific person and clothes, while a Densepose-guided ControlNet is exploited to guarantee consistent prior about body pose across various camera views.
Besides, to avoid the inconsistent multi-view priors from the personalized SD dominating the optimization, DreamVTON introduces a template-based optimization mechanism, which employs mask templates for geometry shape learning and normal/RGB templates for geometry/texture details learning.
Furthermore, for the geometry optimization phase, DreamVTON integrates a normal-style LoRA into personalized SD to enhance normal map generative prior, facilitating smooth geometry modeling.
Extensive experiments show that DreamVTON can generate high-quality 3D Humans with the input person, clothes images, and text prompt, outperforming existing methods.
\footnote{Xiaodan Liang is the corresponding author. This work was done during Zhenyu Xie's internship in ByteDance.}
\end{abstract}

%%
%% The code below is generated by the tool at http://dl.acm.org/ccs.cfm.
%% Please copy and paste the code instead of the example below.
%%


%%
%% Keywords. The author(s) should pick words that accurately describe
%% the work being presented. Separate the keywords with commas.

%% A "teaser" image appears between the author and affiliation
%% information and the body of the document, and typically spans the
% %% page.
% \begin{teaserfigure}
%   \includegraphics[width=\textwidth]{sampleteaser}
%   \caption{Seattle Mariners at Spring Training, 2010.}
%   \Description{Enjoying the baseball game from the third-base
%   seats. Ichiro Suzuki preparing to bat.}
%   \label{fig:teaser}
% \end{teaserfigure}

% \received{20 February 2007}
% \received[revised]{12 March 2009}
% \received[accepted]{5 June 2009}

%%
%% This command processes the author and affiliation and title
%% information and builds the first part of the formatted document.

\begin{teaserfigure}
  \vspace{-4mm}
  \includegraphics[width=1.0\hsize]{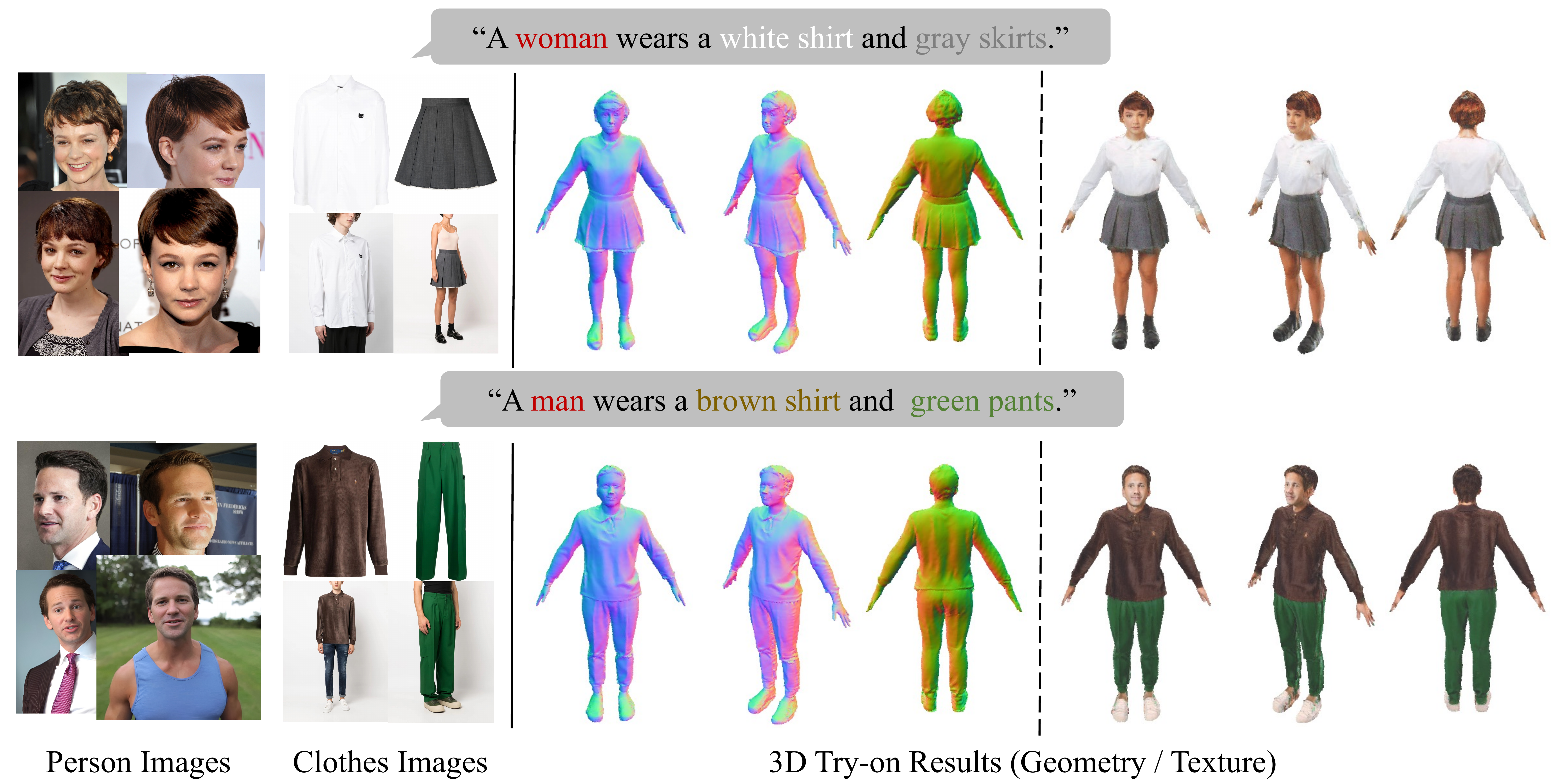}
  \vspace{-6mm}
  \caption{Given a set of person images, clothes images, and a text prompt, our proposed DreamVTON can generate high-quality 3D Humans, wearing customized clothes, keeping the identity and clothes style.}
  % \vspace{4mm}
  \label{fig:teaser}
\end{teaserfigure}

\maketitle

\section{Introduction}\label{sec:intro}
The task of Virtual Try-ON (VTON), to transfer a clothing item onto a specific person, has been explored a lot in recent years due to its promising potential to revolutionize the industry of e-commerce and fashion design.
The image-based 2D solutions take person and clothes images as inputs and achieve virtual try-on via 2D generative models~\cite{NIPS2014_5ca3e9b1,dickstein2015dm,ho2020denoising,rombach2021highresolution}. Although the advanced 2D solutions~\cite{zhenyu2020lgvton,Ge2021PFAFN,zhenyu2021wasvton,zhenyu2021pastagan,dong2022wflow,zaiyu20223dgcl,xie2022pastagan++,he2022fs_vton,xie2023gpvton,gou2023taming,zhang2023warpdiff,zhang2024mmtryon} can synthesize compelling results within particular viewpoint (e.g., front view), they fail to display try-on results for arbitrary observed viewpoint, which is commonly required in the real-world scenarios.
Traditional 3D solutions model the try-on results in the 3D space, thus providing a more comprehensive and attractive perception of clothes fitting.
However, most of these solutions~\cite{peng2012drape,fabian2014scs,gerard2017clothcap,lahner2018deepwrinkles,bhatnagar2019multi} rely upon the 3D scanning equipment or labor-intensive manual annotation, making them much resource-hungry compared with the image-based 2D counterparts.
The pros and cons of the existing 2D and 3D solutions inspire us to rethink whether sculpting the 3D try-on human by simply using the person and clothes images is possible.

\begin{figure}
  \centering
  \includegraphics[width=1.0\hsize]{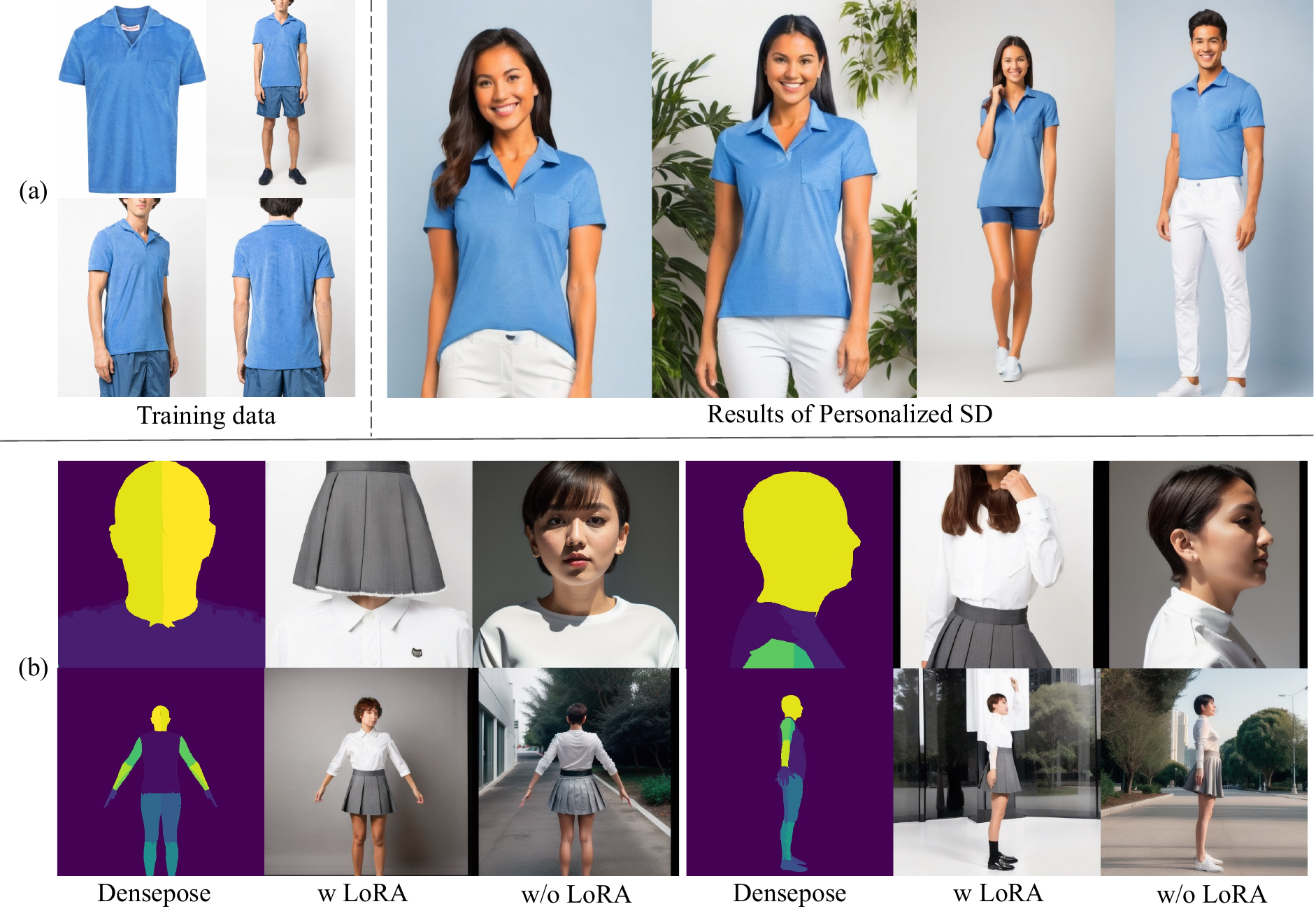}
  \vspace{-6mm}
  \caption{(a) Try-on results of the personalized SD. (b) Visual comparison between results of SD with and without LoRA directly. 
  Using LoRA directly will reduce the capability of SD for multi-view synthesis.
  } 
  \vspace{-6mm}
  \label{fig:lora_results}
\end{figure}

Recently, the extraordinary success of diffusion models~\cite{saharia2022imagen,ramesh2022dalle2,rombach2021highresolution} for text-to-image (T2I) has largely prompted the development of high-quality 3D content generation~\cite{poole2022dreamfusion,raj2023dreambooth3d,lin2023magic3d,chen2023fantasia3d,wang2023prolificdreamer,shi2023mvdream}, whose optimization of the 3D representation is guided by 2D generative priors from the pre-trained T2I diffusion model (e.g., StableDiffusion(SD)~\cite{rombach2021highresolution}) by using Score Distillation Sampling (SDS) loss~\cite{poole2022dreamfusion}. 
Previous 3D human generation works explore this diffusion-based 3D generation framework to sculpt a 3D human according to textual descriptions~\cite{jiang2023avatarcraft,kolotouros2023dreamhuman,zhang2023avatarverse,huang2023humannorm,cao2023dreamavatar,liao2023tada,huang2023dreamwaltz} or reference human images~\cite{zeng2023avatarbooth,huang2024tech}. 
Despite the significant advancement in high-quality 3D modeling, these methods can not be directly adapted to 3D virtual try-on, because they neither take the clothing items as input nor consider the clothing manipulation during the 3D modeling procedure.
Some diffusion-based methods have explored the potential of employing lightweight personalized modules (i.e., LoRA~\cite{hu2022lora}) in SD for 2D virtual try-on. As shown in Figure~\ref{fig:lora_results} (a), by using several clothes images for LoRA fine-tuning, the personalized SD can generate photo-realistic fashion models wearing specific clothes.
Considering the benefits of 2D and 3D generation, it is straightforward to achieve 3D virtual try-on by integrating the personalized SD with the diffusion-based 3D generation framework.
% in which the personalized SD is trained by using person and clothes images and can provide the generative priors about specific person and clothes for the 3D optimization procedure.

However, introducing LoRA into SD would degrade the model's capability for multi-view generation, since it is trained using rare images within limited viewpoints. 
As shown in Figure~\ref{fig:lora_results} (b), for some observed viewpoints, given the same prompt, integrating SD with LoRA would generate wrong results or directly crash, while SD without LoRA can generate realistic results conforming to the input densepose~\cite{PE_facebook2018densepose}.
The degraded ability for multi-view synthesis results in inconsistent generative priors across various viewpoints and further influences the 3D optimization procedure, leading to coarse 3D geometry and blurred texture.
Therefore, it is non-trivial to integrate the personalized SD into a diffusion-based 3D generation framework for image-based 3D virtual try-on.

To target the challenges, we propose a novel diffusion-based 3D human generation framework, named DreamVTON, to sculpt the 3D human by simply taking several person images, clothes images, and a text prompt as inputs (see Figure~\ref{fig:teaser}).
Specifically, our DreamVTON inherits the advanced two-stage 3D generation framework~\cite{chen2023fantasia3d,huang2023humannorm,huang2024tech}, the first stage optimizes the DMTet-based~\cite{gao2020deftet,shen2021dmtet} 3D geometry, while second stage optimizes the texture. During the geometry and texture optimization procedures, DreamVTON introduces a multi-concept LoRA to provide generative priors about the specific person and clothes.
Besides, inspired by AvatarVerse~\cite{zhang2023avatarverse}, DreamVTON employs a Densepose-guided ControlNet~\cite{zhang2023adding} to provide consistent priors about body pose across various viewpoints.
To avoid inconsistent generative priors from personalized SD dominating the 3D optimization procedure, DreamVTON proposes a template-based optimization mechanism, which employs mask templates for precise geometry shape learning and normal/RGB templates for precise geometry/texture details learning. The personalized SD generates the RGB templates within several pre-defined viewpoints, while the mask and normal templates are derived from the RGB templates by using the off-the-shelf mask and normal predictors.
Moreover, to enhance the 3D geometry perception of the personalized SD during the geometry optimization procedure, DreamVTON introduces a normal-style LoRA which can facilitate the personalized SD to provide more powerful prior about the normal map, leading to smoother geometry modeling.

Overall, the main contributions can be summarized: 
(1) We propose a diffusion-based 3D virtual try-on framework, named DreamVTON, which employs personalized SD with multi-concept and normal-style LoRAs to provide powerful generative priors for 3D human optimization. 
(2) We jointly exploit the SDS loss and template-based optimization mechanism for high-quality 3D human generation. 
(3) We further introduce a normal-style LoRA into personalized SD for smoother geometry.
(4) Extensive experiments show that DreamVTON can generate realistic 3D try-on results, consistent with the input images, and outperform baseline methods.

\section{Related Works}
\label{sec:related-works}

% \subsection{2D/3D Virtual Try-on}
\noindent\textbf{2D/3D Virtual Try-on.}
Most 2D VTON methods~\cite{bochao2018cpvton,xintong2019clothflow,han2020acgpn,he2022fs_vton,lee2022hrviton,bai2022sdafn,xie2023gpvton,gou2023taming} employ a two-stage framework to process garment deformation and try-on generation separately, in which the former uses the Thin Plate Splines~\cite{bookstein1989tps} (TPS) or flow-based~\cite{TinghuiZhou2016ViewSB} network to model geometry deformation, while the latter employs generative models like Generative Adversarial Network~\cite{NIPS2014_5ca3e9b1} or Diffusion Model~\cite{rombach2021highresolution} to synthesize the try-on results. 
% Apart from the two-stage methods, 
Some advanced works~\cite{TryonDiffusion,zhang2024mmtryon} introduce an implicit warping mechanism and processes clothes warping and try-on generation within a single diffusion network.
% Although the 2D VTON solutions are resource-efficient (only using image data) and can generate compelling results, they fail to present results for arbitrary viewpoints, which is commonly required in real scenarios.
Traditional 3D VTON methods~\cite{10.1145/566654.566623,lahner2018deepwrinkles,peng2012drape,fabian2014scs,gerard2017clothcap} relies on the 3D scan equipment or cloth simulation to generate geometric representations of high precision. Learning-based methods~\cite{bharat2019multigarment,chaitanya2020tailornet,aymen2020pixsurf,zhu2020deepfashion3d} employ differentiable rendering to dress the SMPL~\cite{SMPL:2015} model with desired garment mesh. M3D-VTON~\cite{zhao2021m3d} proposes a depth-based 3D VTON framework to lift the 2D results to 3D.
% which significantly enhances the geometric details.
Differently, we handle image-based 3D VTON by using the generative model, which can integrate the complementary advantages of 2D and 3D VTON.
% (i.e., resource-efficient and with impressive presentation).

% \subsection{Diffusion-based 3D Human Generation}
\noindent\textbf{Diffusion-based 3D Human Generation.}
Diffusion-based 3D human generation methods\cite{jiang2023avatarcraft,zeng2023avatarbooth,cao2023dreamavatar} aim to generate 3D humans, using text prompts or reference images as input.
% with the desired shape, pose, and color from given text prompts or images. 
% Generally, 
They apply SDS-based optimization\cite{poole2022dreamfusion} to progressively generate 3D humans from initial shape often parameterized by SMPL\cite{SMPL:2015}. 
TADA\cite{liao2023tada} and TeCH\cite{huang2024tech} deploys SMPL-X\cite{pavlakos2019smplxexpressive} expressing 3D human with more detail.
Pose-aware neural human representation imGHUM\cite{alldieck2021imghum} used to generate the human body in DreamHuman\cite{kolotouros2023dreamhuman}.
AvatarBooth\cite{zeng2023avatarbooth} employs DreamBooth\cite{ruiz2023dreambooth} to inject specific identity information into SD, enhancing identity consistency in the personalized 3D human body generation process.
DreamWaltz\cite{huang2023dreamwaltz} and AvatarVerse\cite{zhang2023avatarverse} leverage Pose ControlNet\cite{zhang2023adding} to obtain detailed human body models. HumanNorm~\cite{huang2023humannorm} introduces a normal-aligned diffusion model that allows for custom identities and poses using normal maps in specific regions. Our DreamVTON is a pose-aware 3D VTON pipeline that keeps face identity and clothes style.

% \subsection{Personalized Diffusion Model}
\noindent\textbf{Personalized Diffusion Model.}
Dreambooth\cite{ruiz2023dreambooth} proposes fine-tuning the network using a small set of subject-specific images, which learns specific objectives of the object with a unique identifier.
Textual inversion\cite{gal2022textual} achieves efficient personalization by optimizing text embeddings, which is used to guide the creation of personalized images during inference. SVDiff\cite{han2023svdiff} introduces an innovative approach by optimizing the singular values of weight matrices within the model. Custom Diffusion\cite{kumari2022customdiffusion} focuses on fine-tuning the key and value projection matrices of cross-attention layers, and can jointly train for multiple concepts or combine multiple fine-tuned models through closed-form constrained optimization. LoRA\cite{hu2021lora} introduces novel styles or concepts into pre-trained text-to-image models by optimizing low-rank approximations of weight residuals. 
However, DreamVTON addresses the 3D personalized challenge, enabling the generation of diverse clothes while preserving identity.

\section{Methodology}

\begin{figure*}
  \centering
  \includegraphics[width=1.0\hsize]{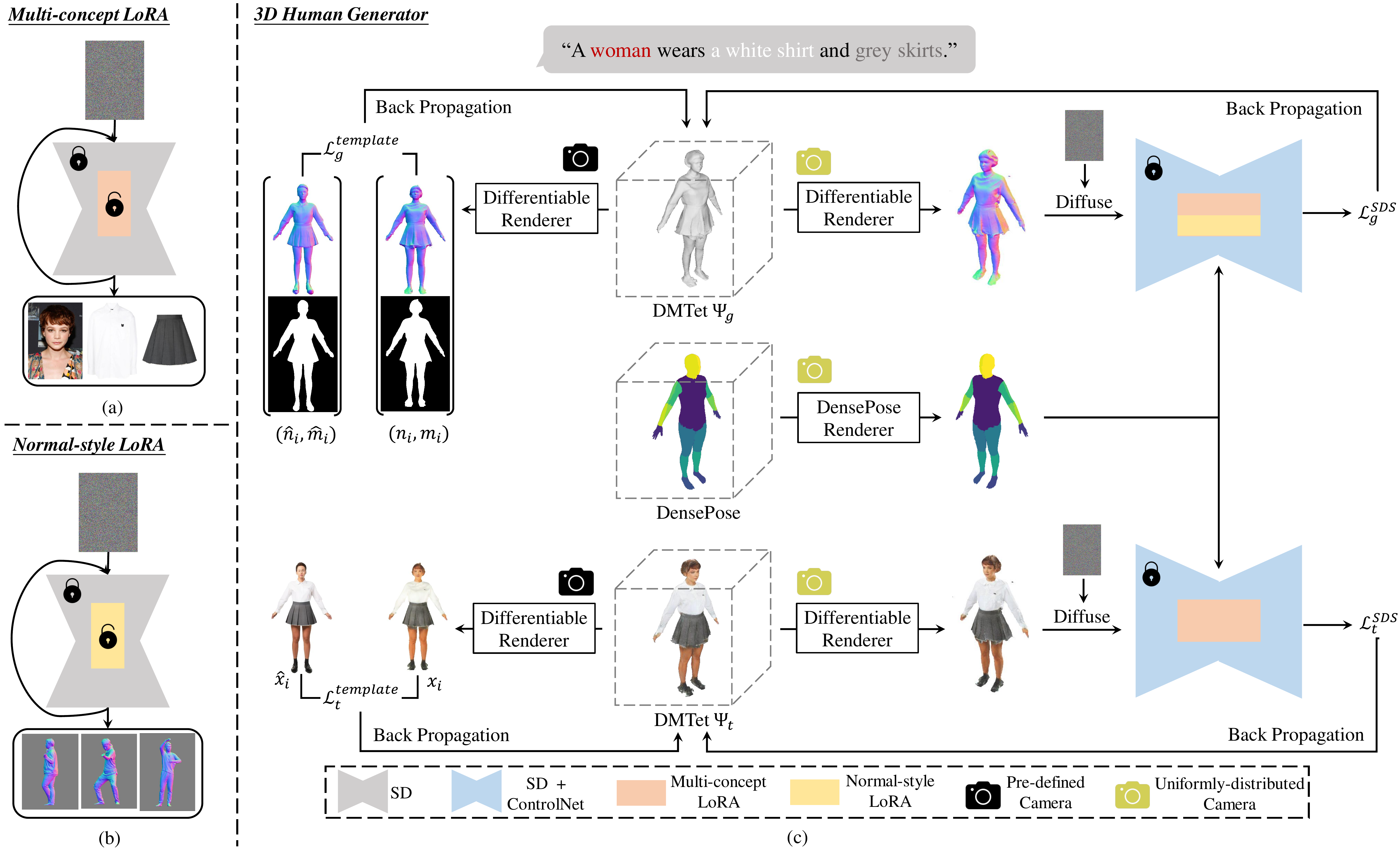}
  \vspace{-6mm}
  \caption{\textbf{Method Overview}. DreamVTON can generate realistic 3D humans given person images, clothes images, and a text prompt. We disentangle the 3D try-on into geometry and appearance learning and design a Multi-concept LoRA and a Normal-style LoRA. Furthermore, we employ a templated-based optimization to achieve high-quality geometry and detailed texture.} 
  \vspace{-4mm}
  \label{fig:framework}
\end{figure*}

The image-based 3D Virtual Try-ON aims to sculpt the 3D digital human using several images of a specific person and clothes items.
To achieve this, we propose DreamVTON, a personalized 3D human generation framework (Sec.~\ref{sec:framework}) that collaboratively employs multi-concept LoRA and Densepose-guided ControlNet to provide the particular generative priors for the 3D optimization procedures of geometry and texture.
To avoid the inferior generative priors from personalized modules (i.e., multi-concept LoRA) dominating the optimization procedure, DreamVTON employs a template-based optimization mechanism (Sec.~\ref{sec:template-based}) to facilitate realistic geometry and texture modeling.
Besides, 
% during the geometry optimization procedure, 
to further enhance the perception of the 3D geometry, DreamVTON introduces a normal-style LoRA (Sec.~\ref{sec:normal-lora}) into the personalized SD.
An overview of DreamVTON is displayed in Figure~\ref{fig:framework}.

\subsection{Personalized 3D Human Generation}\label{sec:framework}
\textbf{Two-stage 3D generation framework.}
To efficiently model high-quality 3D try-on digital human, our DreamVTON inherits the advanced two-stage 3D human generation framework~\cite{chen2023fantasia3d,huang2024tech,huang2023humannorm}, in which the 3D geometry and texture are optimized separately by using Score Distillation Sampling (SDS)~\cite{poole2022dreamfusion} to distill the generative priors from the pre-trained Stable Diffusion (SD)~\cite{rombach2021highresolution} $\epsilon_{\phi}$.

For geometry modeling, DreamVTON utilizes a MLP network $\Psi_{\mathrm{g}}$ to parameterize the DMTet-based~\cite{gao2020deftet,shen2021dmtet} geometry representation $(V_T, T)$, 
% which contains 3D vertices $V_T$ in the deformable tetrahedral grid $T$. 
in which $\Psi_{\mathrm{g}}$ is trained to predict the Signed Distance Function (SDF) value $s_i$ and the deformation offset $\triangle v_i$ for each vertex $v_i \in V_T$ in tetrahedral grid $T$.
During training, DreamVTON first employs an initialization phase to fit $T$ onto an A-pose SMPL~\cite{SMPL:2015} mesh $M_{smpl}$, by using the following object function:
\begin{equation}
    \mathcal{L}_{g}^{\mathrm{init}}=\sum_{p_i \in \mathbf{P}}\left\|s\left(p_i ; \Psi_{\mathrm{g}}\right)-S D F\left(p_i\right)\right\|_2^2,
\end{equation}
where $\mathbf{P}$ is a point set randomly sampled around the surface of $M_{smpl}$, and $S D F\left(p_i\right)$ is the pre-calculated SDF value.
Then, DreamVTON employs the SDS-based optimization mechanism to sculpt the geometry details. To be specific, DreamVTON conducts differentiable rendering onto DMTet mesh to obtain a normal map $\mathbf{n}$, which will then be passed to the pre-trained SD to calculate the normal map SDS loss as follows:
\begin{equation}
\mathcal{L}_{g}^{\mathrm{SDS}}=\mathbb{E}\left[w(\mathbf{t})\left(\epsilon_\phi\left(\mathbf{z_t^{n}} ; \mathbf{c_n}, \mathbf{t}\right)-\epsilon\right) \frac{\partial \mathbf{n}}{\partial \psi_g} \frac{\partial \mathbf{z_t^{n}}}{\partial \mathbf{n}}\right],
\label{eq:geometrysds}
\end{equation}
where $\mathbf{z_t^{n}}$ is the latent code of $\mathbf{n}$ with $\mathbf{t}$-step noising, $\mathbf{c_n}$ is the embedding of normal map prompt extracted by CLIP~\cite{nichol2021clip}, and $\psi_g$ is the parameters of $\Psi_{\mathrm{g}}$.

For texture modeling, DreamVTON utilizes another MLP network $\Psi_{\mathrm{t}}$ to parameterize the material model and uses the Physically-Based Rendering derived from Fantasia3D~\cite{chen2023fantasia3d} to obtain the rendered RGB image $\mathbf{x}$.
During training, DreamVTON feeds $\mathbf{x}$ into pre-trained SD to calculate the image SDS loss as follows:
\begin{equation}
\mathcal{L}_{t}^{\mathrm{SDS}}=\mathbb{E}\left[w(\mathbf{t})\left(\epsilon_\phi\left(\mathbf{z_t^{x}} ; \mathbf{c_x}, \mathbf{t}\right)-\epsilon\right) \frac{\partial \mathbf{x}}{\partial \psi_t} \frac{\partial \mathbf{z_t^{x}}}{\partial \mathbf{x}}\right],
\label{eq:texturesds}
\end{equation}
where $\mathbf{z_t^{x}}$ is the latent code of $\mathbf{x}$, $\mathbf{c_x}$ is the embedding of image prompt, and $\psi_t$ is the parameters of $\Psi_{\mathrm{t}}$.

\begin{figure}
  \centering
  \includegraphics[width=1.0\hsize]{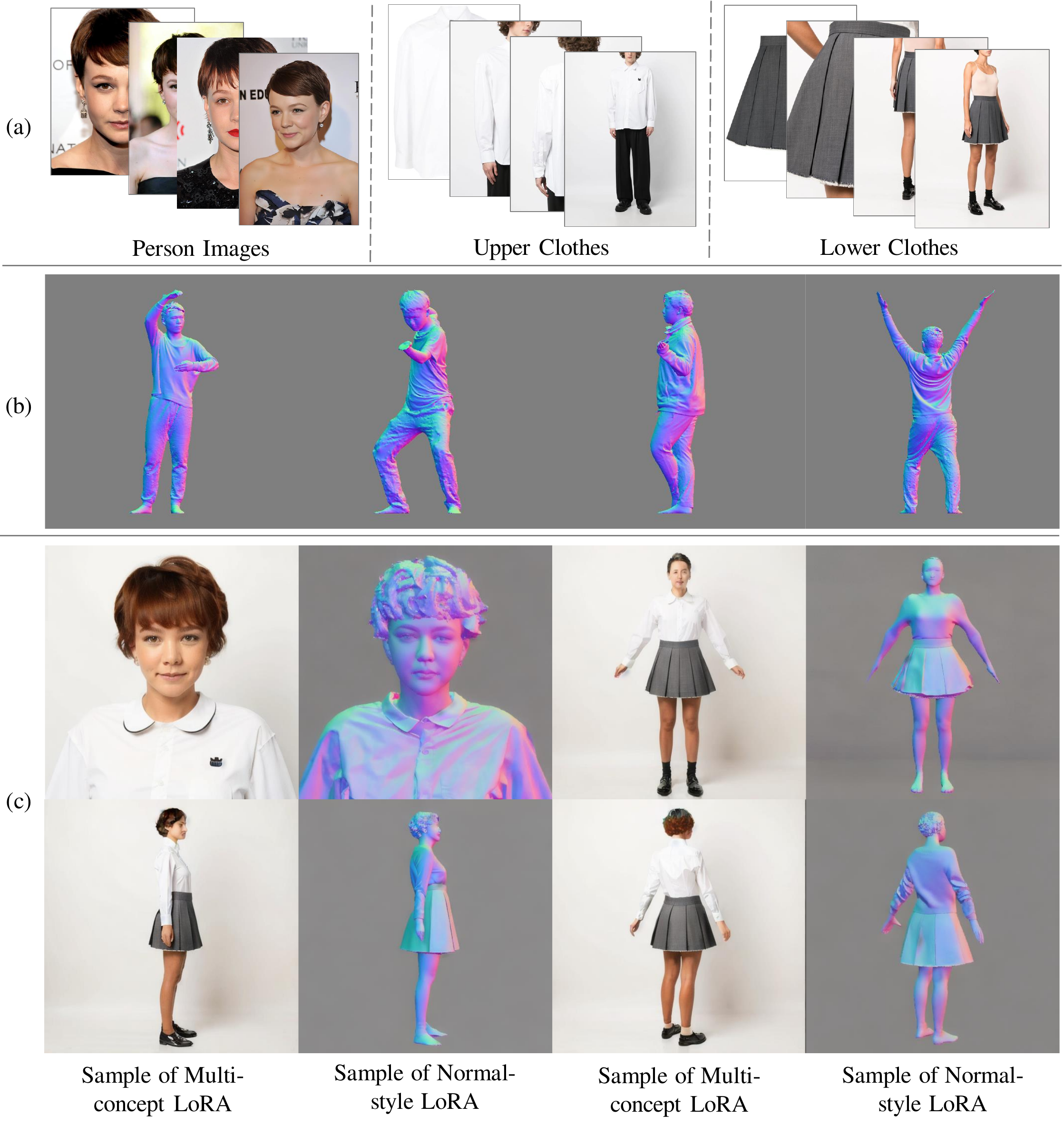}
  \vspace{-6mm}
  \caption{(a) Sample of training data for multi-concept LoRA. (b) Sample of training data for normal-style LoRA. (c) Sample results of multi-concept LoRA and Normal-style LoRA.} 
  \vspace{-4mm}
  \label{fig:normallora_results}
\end{figure}

\textbf{Personalized SD for image-based 3D VTON.}
Although existing methods~\cite{huang2024tech,huang2023humannorm} based on the above two-stage framework can obtain high-quality 3D digital human, they can not be adapted to image-based 3D VTON, because they are incapable of handling clothes inputs.
To address this problem, DreamVTON introduces a multi-concept LoRA to inject the knowledge of the specific person and clothes into pre-trained SD, which will provide the generative priors of clothes and person for 3D optimization.
The multi-concept LoRA is trained by jointly using person images and clothes images. Person images and clothes images separately provide the identity and clothes information for virtual try-on. 
As shown in Figure~\ref{fig:normallora_results} (a), the person images contain several person images of the same person, while the clothes image contains in-shop clothes and fashion models wearing the particular clothes.
As for text prompts, DreamVTON employs the visual-language model BLIP~\cite{li2022blip} to generate captions for each training image.
During the inference stage, the text prompt is constructed by extracting the principal concept of each image set, such as \textit{"a woman wears a white shirt and grey skirt."}
We display additional examples of the text prompts used for training and inference in the supplementary material.
Besides, inspired by AvatarVerse~\cite{zhang2023avatarverse}, DreamVTON exploits a Densepose-guided ControlNet~\cite{zhang2023adding} to provide consistent generative priors about body pose across various viewpoints. 
Therefore, by employing multi-concept LoRA and Densepose-guided ControlNet, the pre-trained SD item in Eq.~\ref{eq:geometrysds} and Eq.~\ref{eq:texturesds} should be modified to $\Tilde{\epsilon}_\phi\left(\mathbf{z_t^{n}}; \mathbf{c_n}, \mathbf{t}, \mathbf{p}\right)$, where $\Tilde{\epsilon}_\phi$ refers to SD with LoRA and ControlNet, while $\mathbf{p}$ refers to DensePose rendered from a SMPL mesh within current camera pose and translation.

\subsection{Template-based 3D Optimization Mechanism}\label{sec:template-based}
SDS-based 3D optimization mechanism commonly samples camera poses uniformly distributed in the 3D space, which enables the generative model (e.g., SD) to provide generative priors from as many viewpoints as possible, and thus facilitates comprehensive 3D optimization.
However, when employing the multi-concept LoRA into SD, the capability of multi-view synthesis largely degrades (as shown in Figure~\ref{fig:lora_results}), since LoRA is trained by using rare images within limited viewpoints. 
The degradation of personalized SD in multi-view synthesis results in inconsistent generative priors across various viewpoints, which is detrimental to 3D optimization.
Therefore, simply using the SDS loss from personalized SD would not be an optimal optimization method for image-based 3D VTON.

To prevent inconsistent generative priors across multiple views from dominating the optimization of 3D models, our DreamVTON introduces a template-based optimization mechanism to facilitate precise geometry and texture learning.
Specifically, DreamVTON first employs personalized SD (i.e., with multi-concept LoRA and Densepose-guided ControlNet) to generate RGB results within $N$ pre-defined viewpoints, which can synthesize realistic results.
The generated results are regarded as the RGB templates $\{\mathbf{\hat{x}_i}\}_{i=1}^N$, which will then be passed into the off-the-shelf parsing predictor~\cite{Gong2019Graphonomy} and normal map predictor~\cite{xiu2022icon} to obtain the mask templates $\{\mathbf{\hat{m}_i}\}_{i=1}^N$ and normal templates $\{\mathbf{\hat{n}_i}\}_{i=1}^N$.

During geometry learning, to guarantee DMTet is optimized into the correct geometry shape, DreamVTON calculate Mean Square Error (MSE) loss $\mathcal{L}_g^{m}$ between the template masks $\{\mathbf{\hat{m}_i}\}_{i=1}^N$ and their corresponding rendered masks $\{\mathbf{m_i}\}_{i=1}^N$ (rendered under the same camera poses with those of templates), which can be formulated as follows:
\begin{equation}
    \mathcal{L}_g^{m} = \sum_{i=1}^N\left\|\mathbf{m_i} - \mathbf{\hat{m}_i}  \right\|_2^2,
\end{equation}
Besides, to facilitate learning of geometry detail, DreamVTON introduces reconstruction losses between $\{\mathbf{\hat{n}_i}\}_{i=1}^N$ and their corresponding rendered normal maps $\{\mathbf{n_i}\}_{i=1}^N$, which consist of a MSE loss $L_g^{mse}$ and a perceptual loss~\cite{johnson2016perceptual} $L_g^{per}$, 
and can be formulated as follows:
\begin{equation}
    \mathcal{L}_g^{mse} = \sum_{i=1}^N\left\|\mathbf{n_i} - \mathbf{\hat{n}_i}  \right\|_2^2,
\end{equation}
\begin{equation}
    \mathcal{L}_g^{per} = \sum_{i=1}^N \sum_{j=1}^5 \lambda_j \left\| \gamma_j(\mathbf{n_i}) - \gamma_j(\mathbf{\hat{n}_i})  \right\|_1,
\end{equation}
where $\gamma_j$ denotas the $j$-th feature map in a pre-trained VGG~\cite{DBLP:journals/corr/SimonyanZ14a} network.
Similarly, during texture learning, to facilitate learning of texture detail, DreamVTON exploits the MSE loss $L_t^{mse}$ and perceptual loss $L_t^{per}$ between $\{\mathbf{\hat{x}_i}\}_{i=1}^N$ and their corresponding rendered RGB images $\{\mathbf{x_i}\}_{i=1}^N$.

It is worth noting that, since the normal and RGB templates are derived from the same generated results, the geometry and texture details on each normal-RGB template pair (i.e., templates rendered under the same camera pose) are strictly aligned. By using the detail-aligned templates for geometry and texture learning, DreamVTON is capable of generating geometry-texture consistent 3D human.

By jointly using the SDS-based and template-based optimization mechanisms, the overall object functions for geometry and texture optimization can be formulated as follows:
\begin{equation}
    \mathcal{L}_g = \mathcal{L}_{g}^{\mathrm{SDS}} + \lambda_g^m \mathcal{L}_g^{m} + \lambda_g^{mse} \mathcal{L}_g^{mse} + \lambda_g^{per} \mathcal{L}_g^{per},
\end{equation}
\begin{equation}
    \mathcal{L}_t = \mathcal{L}_{t}^{\mathrm{SDS}} + \lambda_t^{mse} \mathcal{L}_t^{mse} + \lambda_t^{per} \mathcal{L}_t^{per},
\end{equation}
where $\lambda_g^*$ and $\lambda_t^*$ are the trade-off hyperparameters.

\begin{figure*}[t]
  \centering
  \includegraphics[width=1.0\hsize]{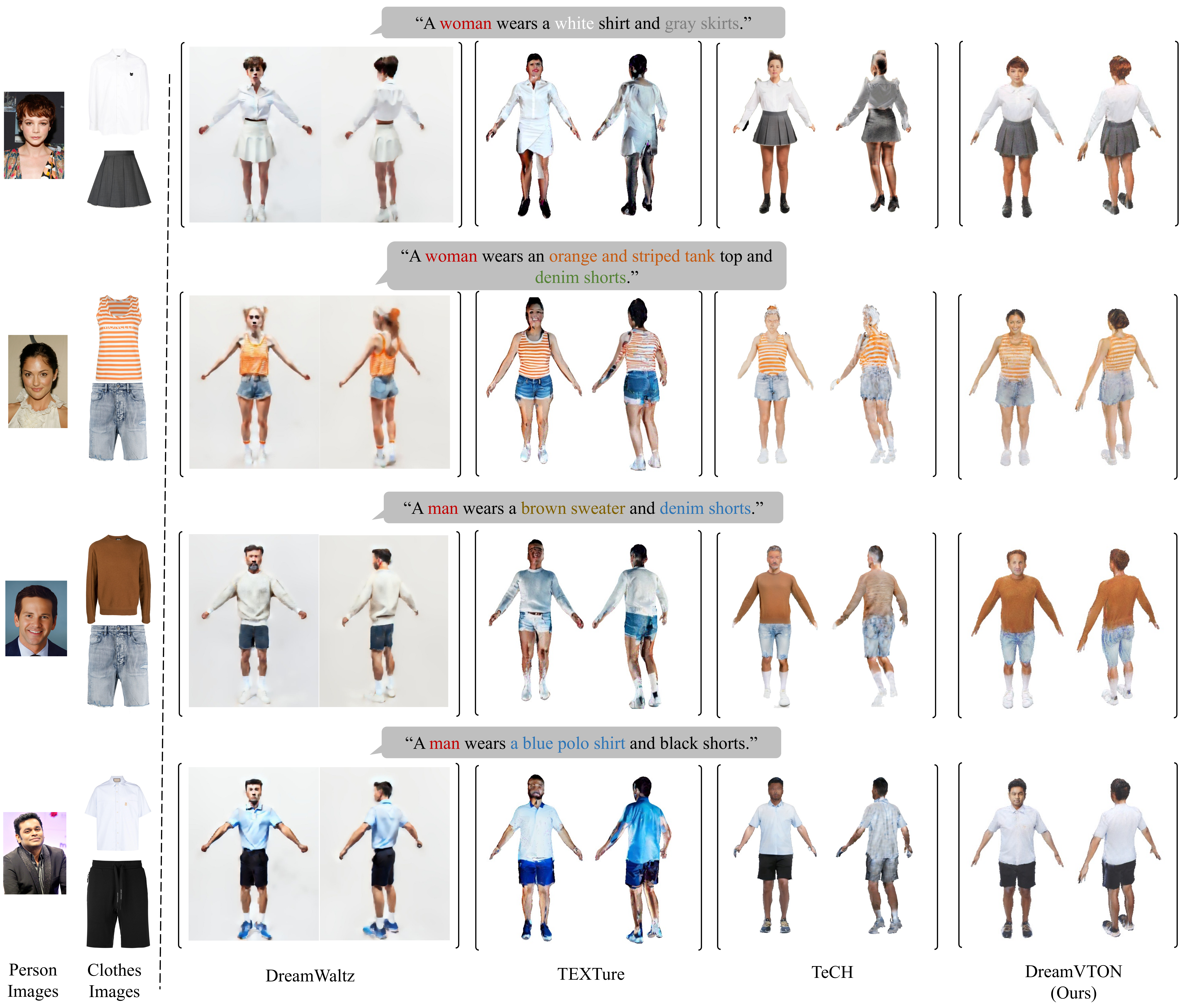}
  \vspace{-4mm}
  \caption{Qualitative Comparisons. Using the same
clothes images, person images, and text prompt as inputs, our method achieves superior results.} 
  \vspace{-4mm}
  \label{fig:vis_comp}
\end{figure*}

\subsection{Normal-style LoRA for Geometry Learning}\label{sec:normal-lora}
During the geometry optimization stage, since DreamVTON employs the rendered normal map to calculate the SDS loss, the personalized SD is designed to provide the normal-style generative prior for geometry optimization.
However, when introducing the multi-concept LoRA, the personalized SD's capability for normal map synthesis degrades a lot, since LoRA is trained by using seldom RGB images.
To address this issue, our DreamVTON introduces normal-style LoRA into personalized SD to enhance the capability for normal map synthesis.
Specifically, the normal-style LoRA is trained with 2,000 text-annotated normal maps, in which the normal maps are rendered from the THUman2.0 dataset~\cite{tao2021function4d}, while the text prompts are extracted by BLIP~\cite{li2022blip}.
Once trained, the normal-style LoRA is integrated into the personalized SD and jointly used for geometry optimization.

As shown in Figure~\ref{fig:normallora_results} (c), given the same prompt \textit{"a woman wears a white shirt and grey skirt, \textbf{with normal map style}."}, SD with multi-concept LoRA can only generate RGB images, while SD with multi-concept and normal-style LoRAs can generate normal-style results, demonstrating that introducing the normal-style LoRA could improve the personalized SD's normal synthesis capability.

\section{Experiments}

We first introduce the implementation details of DreamVTON (Sec.~\ref{sec:implementation}), which contains the dataset description, training configuration, camera sampling strategy, and template selection mechanism. 
Then, we compare DreamVTON with existing 3D human generation methods qualitatively and quantitatively (Sec.~\ref{sec:qualitative}, Sec.~\ref{sec:quantitative}, and Sec.~\ref{sec:user_study}).
Finally, we conduct ablation studies to verify the effectiveness of the proposed modules in DreamVTON (Sec.~\ref{sec:ablation}).

\subsection{Implementation Details}\label{sec:implementation}

\textbf{Dataset Description.}
Since there is no existing dataset tailored for the task of image-based 3D virtual try-on, we collect a new dataset from the internet, which comprises images of 10 various individuals and 33 clothes items (i.e., upper clothes, lower clothes, dresses.). Specifically, most of the individual images are portrait images and each individual contains about 10 portrait images. On the other hand, the clothes images contain in-shop clothes and fashion models wearing particular clothes.
By matching the portrait and clothes images, we can obtain the person-clothes pairs used for the training of image-based 3D virtual try-on. In our experiments, we construct 18 person-clothes pairs, which is then used in our 3D try-on experiments. Some visual examples can be found in Figure~\ref{fig:normallora_results}.

\noindent \textbf{Training configuration.}
The geometry network $\Psi_{\mathrm{g}}$ and texture network $\Psi_{\mathrm{t}}$ are trained 15000 and 3000 iterations, respectively. The training procedure of $\Psi_g$ can be further divided into 2000 iterations SDF initialization phase and 13000 iterations DMTet optimization phase.
During training, the batch size on each GPU is set to 1 and both networks are trained by using AdamW~\cite{loshchilov2019adamw} optimizer. The learning rate for $\Psi_{\mathrm{g}}$ and $\Psi_{\mathrm{t}}$ are set to 1e-3 and 1e-2, respectively.
Both $\Psi_{\mathrm{g}}$ and $\Psi_{\mathrm{t}}$ are trained on one NVIDIA 4090 GPUs.
For each try-on instance, the data preparation and training (excluding the shared Normal-style LoRA) take approximately 90 minutes.

% and the overall optimization procedure costs about 30 minutes for one try-on procedure.

\noindent \textbf{Camera sampling strategy.}
For geometry learning, two type of cameras (i.e., global cameras and local cameras) are employed for optimization, in which the global cameras are located in a position that can cover the full human body, while the local cameras posed to positions that focus on various local regions (i.e., head, upper body, and lower body). Within the local cameras, SD can provide more detailed generative prior for geometry optimization. During training, the sampling probabilities for global cameras and local cameras are set to 0.7 and 0.3, respectively.
For texture learning, only the local cameras are employed, which enhance the learning of texture details.

\noindent \textbf{Selection of geometry/texture templates.}
For geometry learning, we employ eight mask templates $\{\mathbf{\hat{m}_i}\}_{i=1}^8$ and two normal templates $\{\mathbf{\hat{n}_i}\}_{i=1}^2$ for optimization, in which $\{\mathbf{\hat{m}_i}\}_{i=1}^8$ are sampled uniformly around the human body while $\{\mathbf{\hat{n}_i}\}_{i=1}^2$ is composed of the front and back view normal maps. 
For texture learning, we utilize three RGB templates $\{\mathbf{\hat{x}_i}\}_{i=1}^3$, which contain front and back views of full body images and one head image. The head image is used to enhance the texture detail around the face region.

\subsection{Qualitative Results}\label{sec:qualitative}
We compare our DreamVTON with three existing 3D human generation methods, namely DreamWaltz~\cite{huang2023dreamwaltz}, TEXTure~\cite{richardson2023texture}, and TeCH~\cite{huang2024tech}. Since DreamWaltz and TEXTure take as input merely the text prompt, we use the constructed text prompts (used by our DreamVTON) for them. For TeCH, since it receives the text and image inputs, except for the constructed text prompt, we also feed the front-view RGB template (used by our DreamVTON) for it.
Figure~\ref{fig:vis_comp} displays the qualitative comparison of DreamVTON with the baselines. By simply receiving the text prompt as model input, DreamWaltz, and TEXTeure can generate 3D humans with similar clothes types to the input clothes images. However, they fail to preserve the texture detail or clothes color. For example, for the white-tshirt-grey-skirts case in the first case, they fail to generate grey skirts.
TeCH~\cite{huang2024tech} takes both text and images as inputs, thus is capable of preserving the clothes details in front view. However, it fails to generate realistic texture in the back view and face region, 
since it ignores the guidance about the back view and face.
In comparison, DreamVTON can not only preserve the clothes information in arbitrary view but also generate a realistic face, demonstrating that it can outperform the compared methods.

\subsection{Quantitative Comparison}\label{sec:quantitative}
Inspired by HumanNorm~\cite{huang2023humannorm}, we choose CLIP-Similarity~\cite{nichol2021clip} and FID~\cite{heusel2017fid} to evaluate the generated quality of the 3D try-on results, in which FID measures the realism of rendered results, while CLIP-Similarity measures the similarity between the rendered results from arbitrary viewpoint and the particular reference images.
Specifically, for each test case, we first employ the personalized SD (Densepose-based ControlNet + Multi-concept LoRA) to generate eight 2D try-on results, of which the camera viewpoints are uniformly distributed around the human body. 
Then, we employ similar but much denser cameras to render another 100 images from the learned 3D try-on results.
we obtain 100 rendered images from the learned 3D try-on results.
During Calculating FID, we regard the 2D generated images as the ground truth and measure the distribution similarity between the pseudo ground truth (i.e., 2D generated results) and the rendered results. We calculate the FID score five times using different samples and report the results with mean and standard deviation.
During calculating CLIP-similarity, we regard the SD-generated images as reference images and calculate the average CLIP distance between the rendered images and reference images.
As reported in Table~\ref{tab:quantitative}, DreamVTON obtains the lowest FID score, which indicates the images rendered by DreamVTON are most closely aligned to the generated results of the personalized SD. Besides, DreamVTON obtains the highest CLIP-similarity, which further demonstrates the rendered images are more consistent with the reference images.
Both of the reported scores illustrate the superiority of DreamVTON over existing baselines.

\subsection{User Study}\label{sec:user_study}

We evaluate our proposed DreamVTON's performance against other methods using the user study. As reported in Table~\ref{tab:user_study}, a higher score for user evaluation indicates that humans preferred the performance, our proposed DreamVTON outperforms all the compared methods. In particular, our proposed DreamVTON significantly outperforms the compared methods in terms of texture realism and ID fidelity, with 91.5\% and 94\% of users, preferring to choose our model. Regarding the quality of the geometry, 78.8\% of users still prefer to choose our method. Overall, Table~\ref{tab:user_study} demonstrates the effectiveness of our method, which outperforms the other methods on texture, ID fidelity, and geometry, respectively. This means that our proposed DreamVTON can generate more realistic-looking 3D humans that are preferred by users, wearing different clothes with accurate 3D geometry and detailed textures.

\begin{table}[t]
\vspace{-2mm}
\caption{Quantitative Comparisons in the collected datasets.}
\vspace{-4mm}
\def\arraystretch{1.2}
\small
\tabcolsep 2.5pt
\centering
    \begin{tabular}{c|cccc}
    \toprule
        & DreamWaltz & TEXTure & TeCH & DreamVTON \\
        \hline
         FID ($\downarrow$) & \et{170.0}{1.958} & \et{163.5}{2.480} & \et{144.2}{1.757} &  \etb{141.0}{1.262} \\
         CLIP-Similarity ($\uparrow$) & 0.596 & 0.613 & 0.655 &  \textbf{0.665} \\
    \bottomrule
    \end{tabular}
    \label{tab:quantitative}
\vspace{-1mm}
\end{table}

\begin{table}[t]
\vspace{-2mm}
\caption{User study results about the 3D generation quality in terms of Geometry, Texture and ID Fidelity.}
\vspace{-4mm}
\def\arraystretch{1.2}
% \small
\tabcolsep 2.5pt
\centering
    \begin{tabular}{c|cccc}
    \toprule
        Preference($\uparrow$) & DreamWaltz & TEXTure & TeCH & DreamVTON \\
        \hline
         Geometry & 0.5\% &  18.0\% &  2.6\% &  \textbf{78.8\%} \\
         Texture & 1.3\% &  1.3\% &  6.0\% & \textbf{91.5\%} \\
         ID Fidelity &  1.8\% &  1.4\% &  2.9\% &  \textbf{94.0\%} \\
    \bottomrule
    \end{tabular}
    \label{tab:user_study}
\vspace{-1mm}
\end{table}

\begin{figure*}[t]
  \centering
  \includegraphics[width=1.0\hsize]{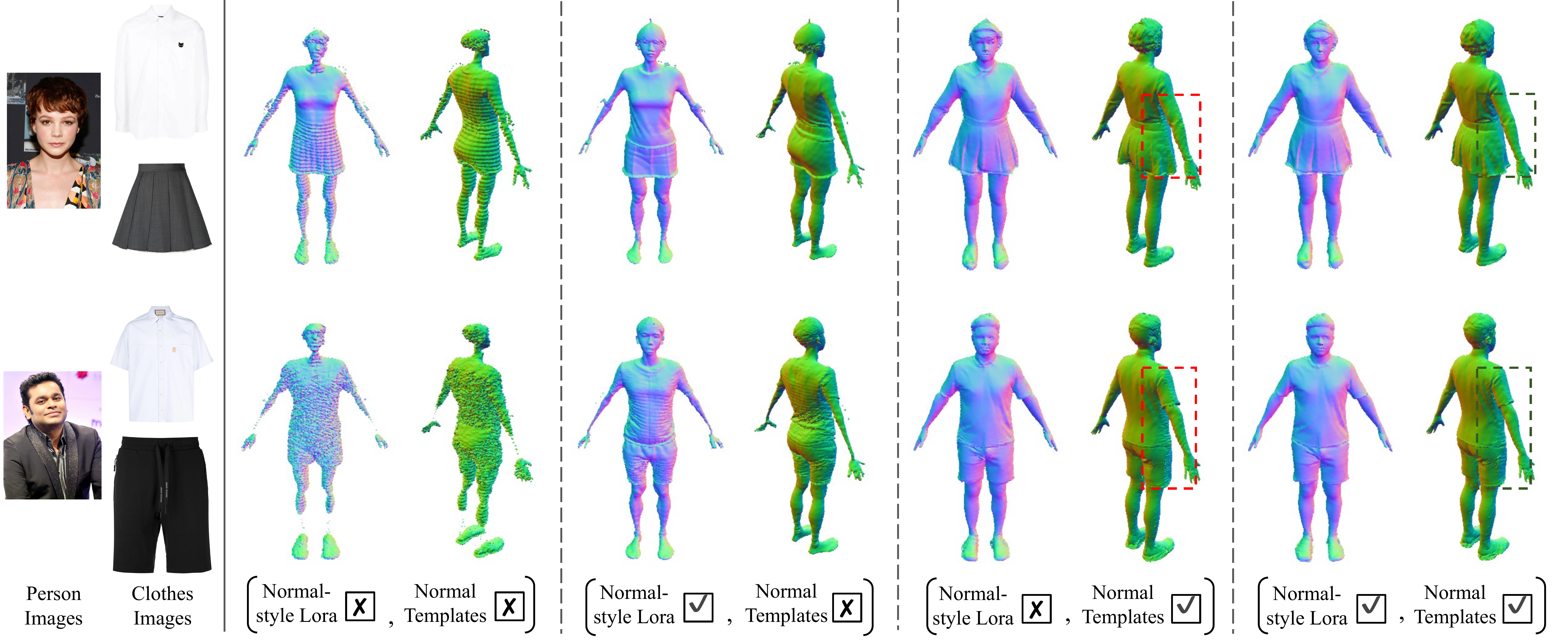}
  \vspace{-6mm}
  \caption{Ablation study for geometry optimization.} 
  \vspace{-2mm}
  \label{fig:abs_geometry}
\end{figure*}

\begin{figure}[t]
  \centering
  \includegraphics[width=1.0\hsize]{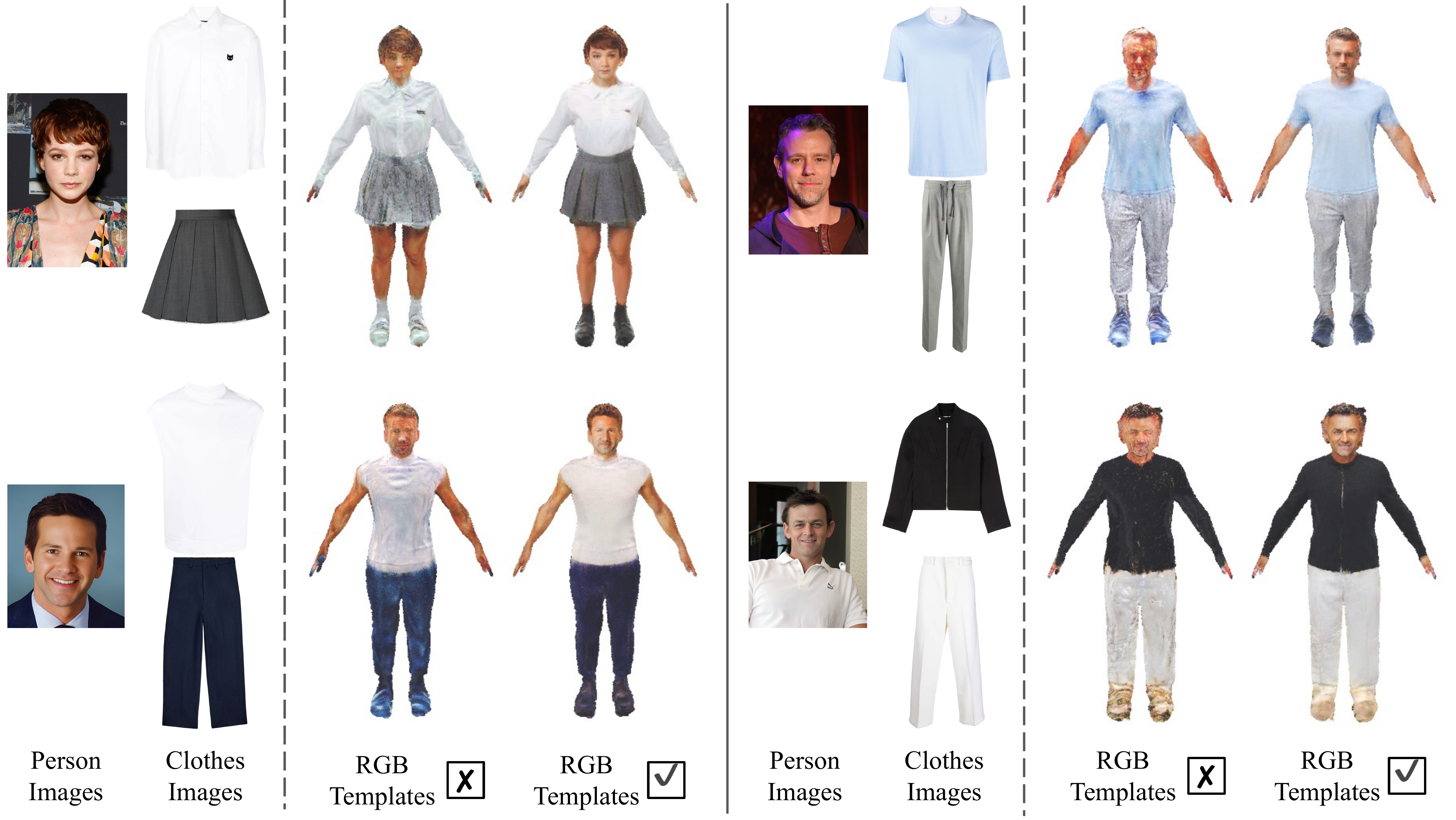}
  \vspace{-4mm}
  \caption{Ablation study for texture optimization.} 
  \vspace{-4mm}
  \label{fig:abs_texture}
\end{figure}

\subsection{Ablation Study}\label{sec:ablation}
We conduct ablation studies to validate the effectiveness of template-based optimization mechanisms and normal-style LoRA for geometry and texture optimization, using visual comparison when excluding the component.

For geometry optimization, as shown in Figure~\ref{fig:abs_geometry}
, without using normal-style LoRA or normal templates as optimization constraints, the surface of the learned geometry is coarse with artifacts. Either adding the normal-style LoRA or using the normal templates in the geometry optimization procedure can smooth the geometry surface. By jointly using normal-style LoRA and normal templates, the geometry surface can be smoother while preserving the clothes characteristics in the input images.

For texture optimization, as shown in Figure~\ref{fig:abs_texture}, without using the RGB templates as optimization constraints, the learned texture is noisy (e.g., unclear face region) and fails to preserve the characteristic of inputs image (e.g., incorrect clothes color).
By using the RGB templates during optimization, DreamVTON can generate 3D humans with high-quality texture and retain the input images' characteristics (i.e., person identity, clothes style).

\section{Limitation}

\begin{figure}[t]
  \centering
  \includegraphics[width=1.0\hsize]{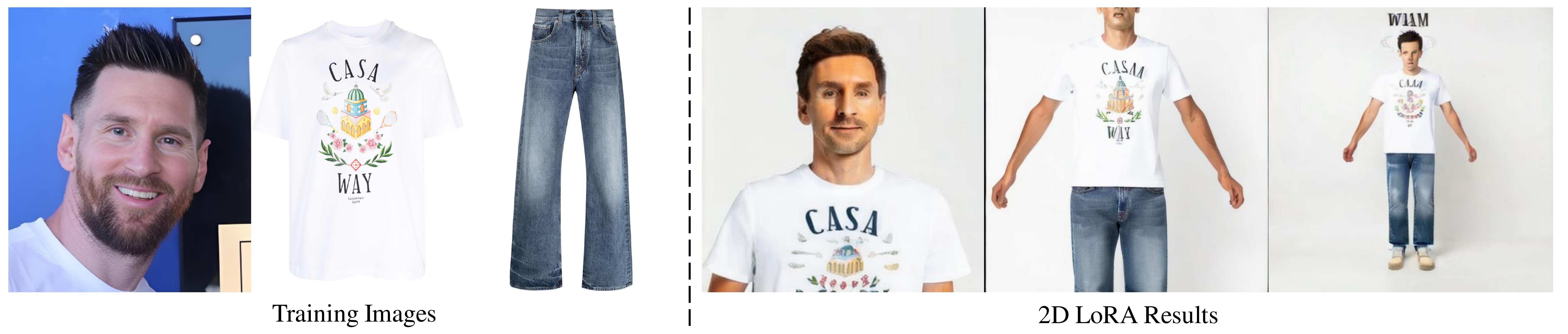}
  \vspace{-8mm}
  \caption{Results of multi-concept LoRA. With complicated logos, LoRA fails to keep logo texture completely.} 
  \vspace{-4mm}
  \label{fig:limitation}
\end{figure}

The texture detail of clothes in DreamVTON's result is derived from the generative priors from multi-concept LoRA. However, existing LoRA fails to preserve the texture detail for complicated logos, (as shown in Figure~\ref{fig:limitation}), thus preventing DreamVTON from generating detailed textures that are completely consistent with input images. 
This problem could be alleviated by balancing the personalization and generalizability of personalized SD, which is widely explored in the diffusion-based models.
% We will make further exploration in the future.

\section{Conclusion}
We propose a new method for 3D virtual try-on task, named DreamVTON, which is capable of generating the 3D human using person images, clothes images, and a text prompt as inputs.
DreamVTON employs an SDS-based framework and disentangles the 3D try-on task as the geometry and texture separately. Specifically, DreamVTON introduces a personalized SD with multi-concept LoRA and Densepose-guided ControlNet to provide powerful pose consistent priors for 3D human optimization. 
DreamVTON designs a templated-based optimization mechanism for precise geometry and texture learning to avoid the degraded multi-view priors from personalized SD. 
In addition, DreamVTON integrates a normal-style LoRA into personalized SD during geometry optimization, which further facilitates smooth and accurate geometry.
Extensive experiments demonstrate that our DreamVTON outperforms existing baselines in terms of geometry and texture model, and can customize high-quality 3D humans with diverse clothes, preserving the person's identity and clothes style.

\clearpage

\section{Acknowledgments}

This work was supported in part by National Science and Technology Major Project (2020AAA0109704), National Science and Technology Ministry Youth Talent Funding No. 2022WRQB002, Guangdong Outstanding Youth Fund (Grant No. 2021B1515020061), Mobility Grant Award under Grant No. M-0461, Shenzhen Science and Technology Program (Grant No. GJHZ20220913142600001), Nansha Key RD Program under Grant No.2022ZD014.

\section{Additional Experiment Details}\label{sec:exp_detail}

\subsection{ Training details of multi-concept LoRA}
For each multi-concept LoRA, the training data consists of about 10/5/5 images of portraits/upper clothes/lower clothes. Captions for each image are extracted using BLIP~\cite{li2022blip}. All images have a resolution of 512x512. We choose Realistic\_Vision\_V5.0 as our basic SD model with the LoRA module integrated into the U-Net and text encoder, setting the LoRA rank to 64. 
During training, the Lion~\cite{chen2023symbolicdiscoveryoptimizationalgorithms} optimizer is employed, and the network is trained with 30 epochs. The learning rates for U-Net and text encoder are set to 1e-4 and 1e-5, respectively.

\subsection{Visual Examples of Training Templates}
\begin{figure}[t]
  \centering
  \includegraphics[width=1.0\hsize]{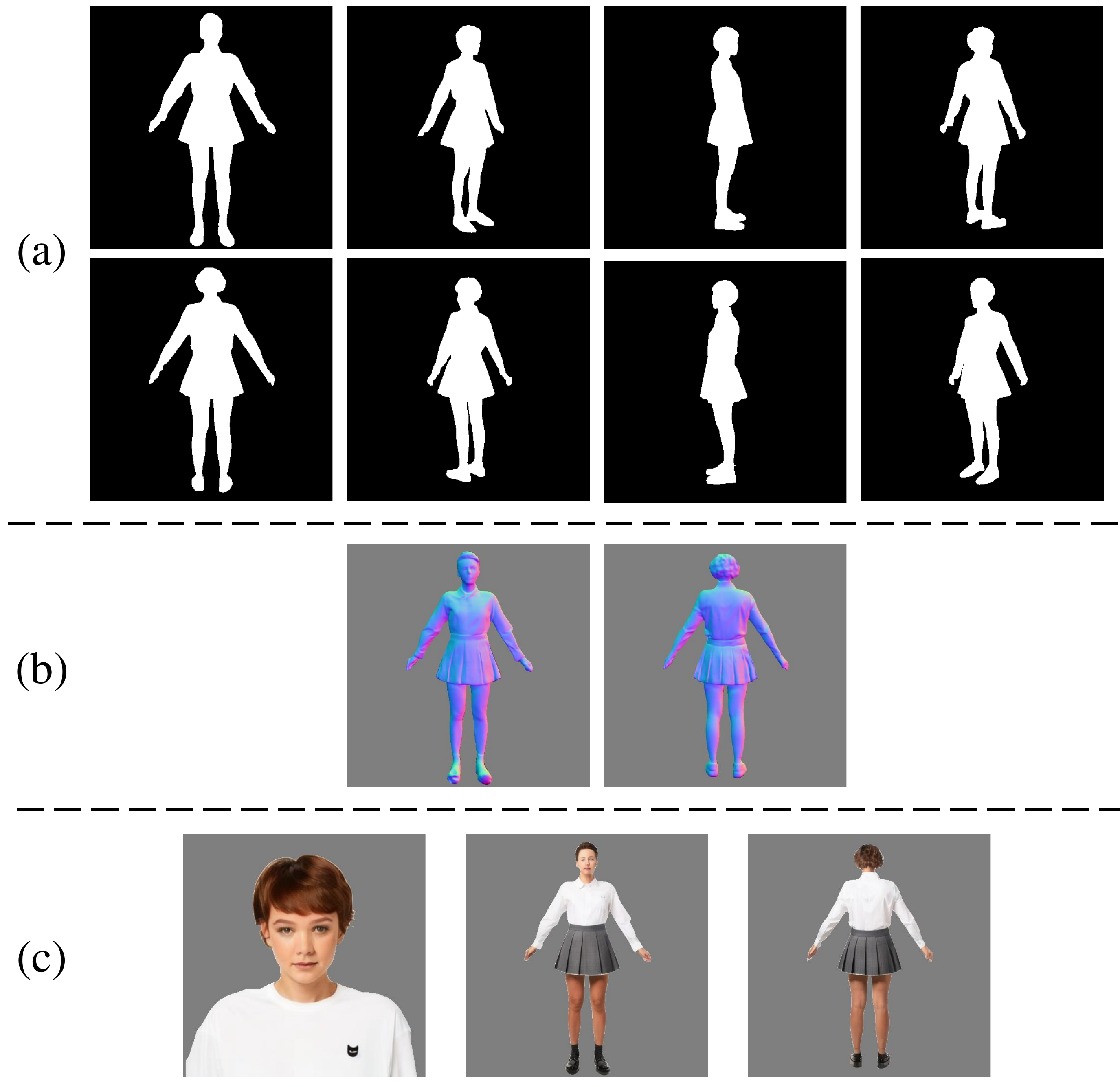}
  % \vspace{-4mm}
  \caption{(a) Examples of mask templates. (b) Examples of normal templates. (c) Examples of RGB templates.} 
  % \vspace{-4mm}
  \label{fig:template}
\end{figure}

As mentioned in Section~\ref{sec:template-based}, our DreamVTON employs various generated template images for geometry and texture optimization.
For geometry optimization, DreamVTON uses eight mask templates $\{\mathbf{\hat{m}_i}\}_{i=1}^8$ (i.e., uniformly distributed views around the human body) to constrain the geometry shape learning, and uses two normal templates $\{\mathbf{\hat{n}_i}\}_{i=1}^2$ (i.e., front view and back view) to enhance the geometry detail learning. Figure~\ref{fig:template} (a) and (b) display examples of mask and normal templates for one try-on use case.
For texture optimization, DreamVTON uses three RGB templates $\{\mathbf{\hat{x}_i}\}_{i=1}^3$ (i.e., front and back full body view, and front face view) to enhance the texture detail learning. Figure~\ref{fig:template} (c) displays examples of RGB templates for one try-on use case.

\subsection{Details of User Study}
\begin{figure*}[t]
  \centering
  \includegraphics[width=1.0\hsize]{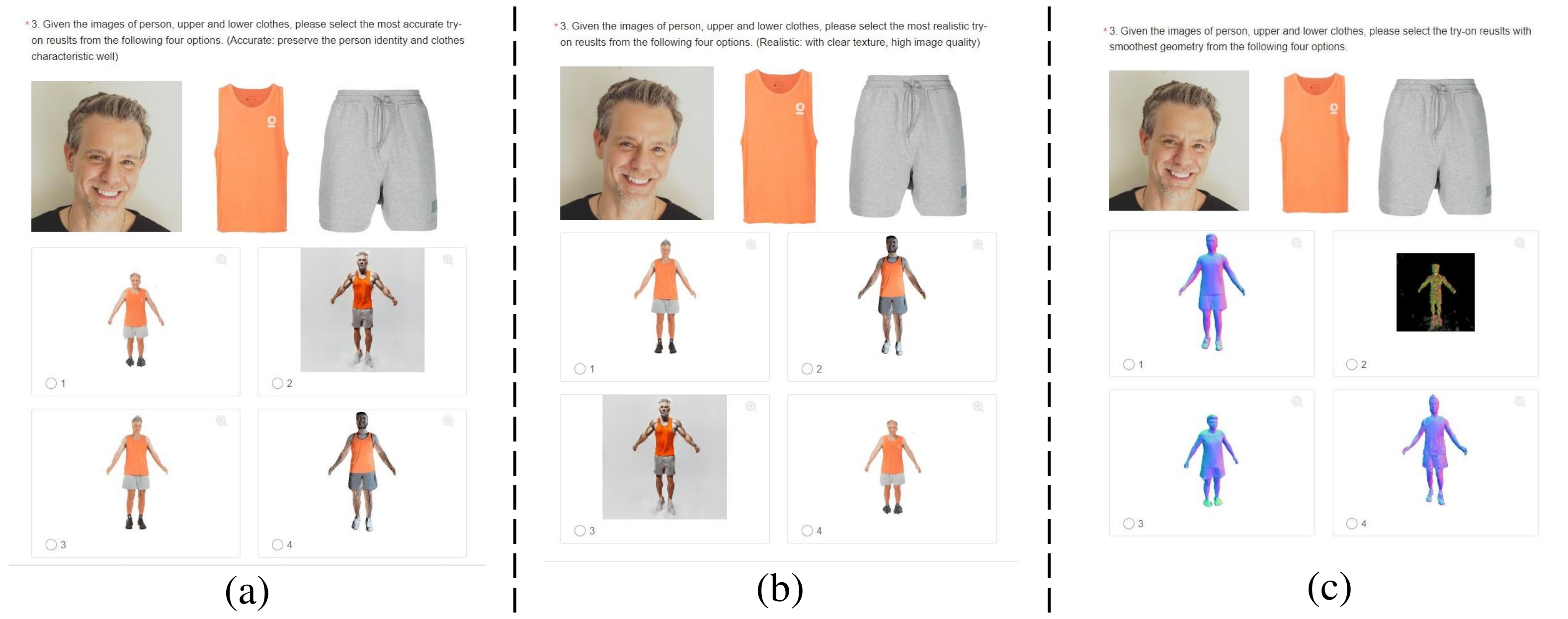}
  % \vspace{-6mm}
  \caption{Interface of the questionnaires used to evaluate \textbf{Accuracy}, \textbf{Realism}, and \textbf{Geometry Smoothness} of the generated 3D try-on results.} 
  % \vspace{-6mm}
  \label{fig:user_study}
\end{figure*}

For user study, we design three questionnaires to separately evaluate the \textbf{Accuracy} (as shown in Figure~\ref{fig:user_study} (a)), \textbf{Realism} (as shown in Figure~\ref{fig:user_study} (b)), and \textbf{Geometry Smoothness} (as shown in Figure~\ref{fig:user_study} (c)) of the 3D try-on results generated by various methods. 
To be specific, \textbf{Accuracy} is used to measure whether the try-on results can preserve a particular person's identity and clothes characteristics, while \textbf{Realism} and \textbf{Geometry Smoothness} are used to evaluate the quality of generated texture and geometry, respectively.

Each questionnaire is composed of 9 assignments, and the amounts of volunteers for these three questionnaires (i.e., (a), (b), (c) in Figure~\ref{fig:user_study}) are 31, 26, and 21, respectively.
For each assignment in the questionnaire, given the person and clothes images, volunteers are asked to select the best 3D try-on result (presented in video format) out of four options, which are generated by our DreamVTON and the other baseline methods (i.e., DreamWaltz~\cite{huang2023dreamwaltz}, TEXTure~\cite{richardson2023texture}, TeCH~\cite{huang2024tech}). Besides, the order of the generated results in each assignment is randomly
shuffled. Figure~\ref{fig:user_study} shows the interface for each questionnaire. For analysis of the user study results, please refer to Section~\ref{sec:user_study}.

\section{Additional Experiment Results}\label{sec:exp_result}

\subsection{Body Shape Controlling Results}
\begin{figure}
  \centering
  \includegraphics[width=1.0\hsize]{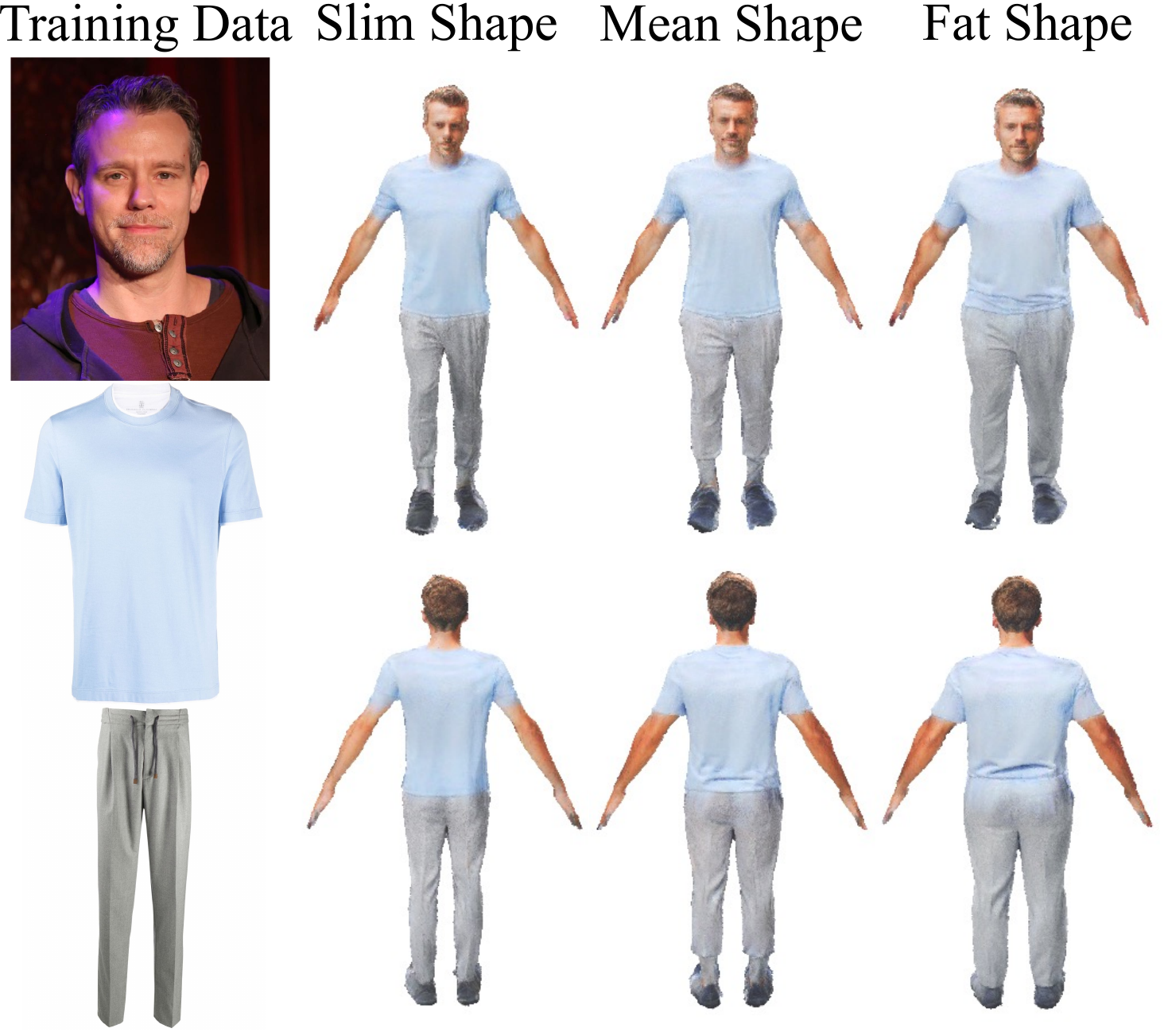}
  \vspace{-6mm}
  \caption{Results for body shape controlling.
  } 
  \vspace{-6mm}
  \label{fig:body_shape}
\end{figure}

To specify the body shape in DreamVTON, we can simply include a full-/half-body image in the training data, from which the body shape information can be easily obtained by using the off-the-shelf 3D pose estimator. 
Subsequently, DreamVTON employs SMPL with a specific body shape as the initial mesh and sculpts the 3D try-on result based on it. 
As displayed in Figure~\ref{fig:body_shape}, using different initial SMPLs, DreamVTON is capable of generating 3D results wearing the same clothes with different body shapes.

\subsection{More Visual Comparison Results}
\begin{figure*}[t]
  \centering
  \includegraphics[width=1.0\hsize]{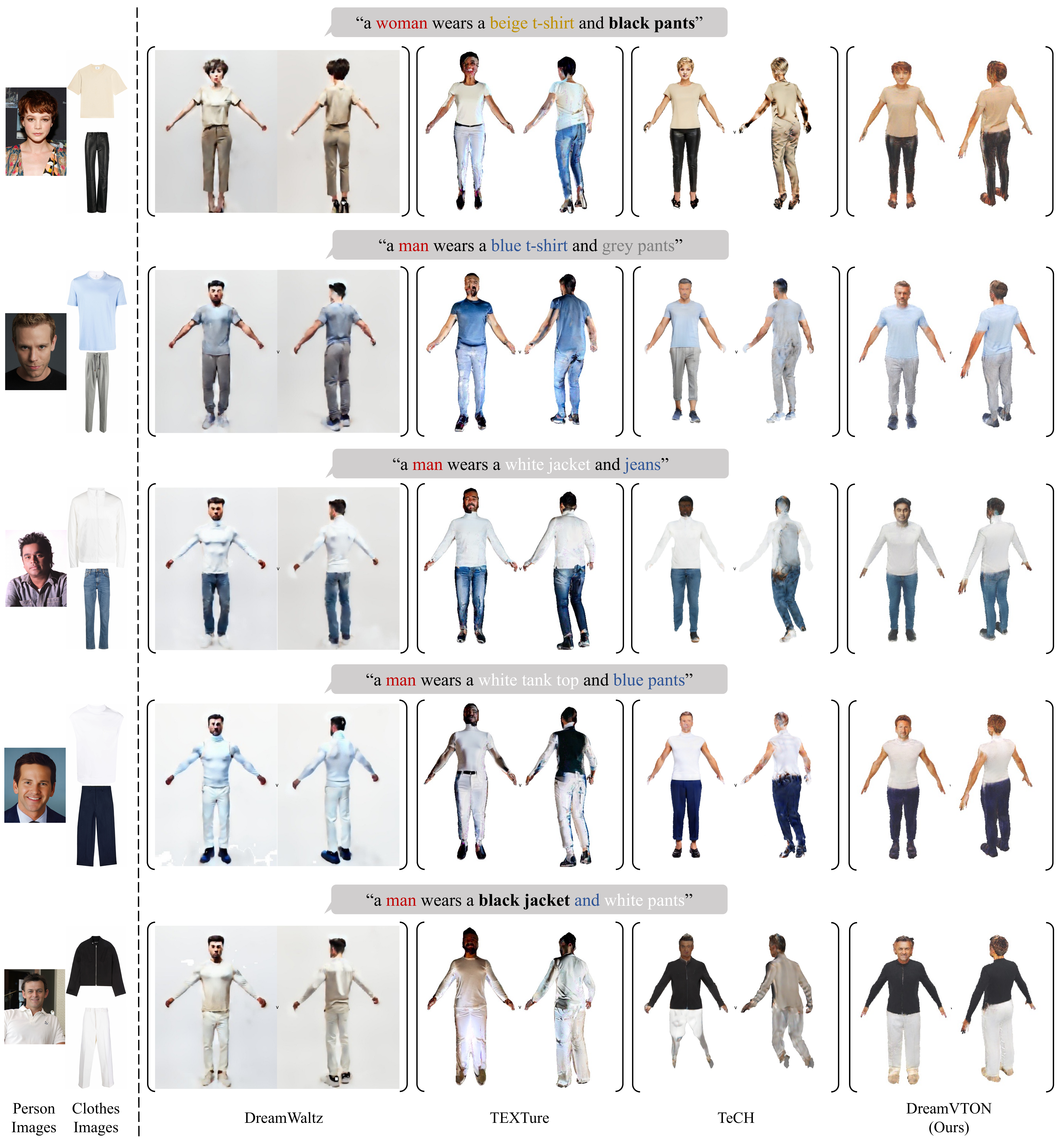}
  % \vspace{-6mm}
  \caption{Qualitative Comparisons. Using the same
clothes, person image, and prompt as inputs, our method achieves superior results.} 
  % \vspace{-6mm}
  \label{fig:vis_comp_v1}
\end{figure*}

\begin{figure*}[t]
  \centering
  \includegraphics[width=1.0\hsize]{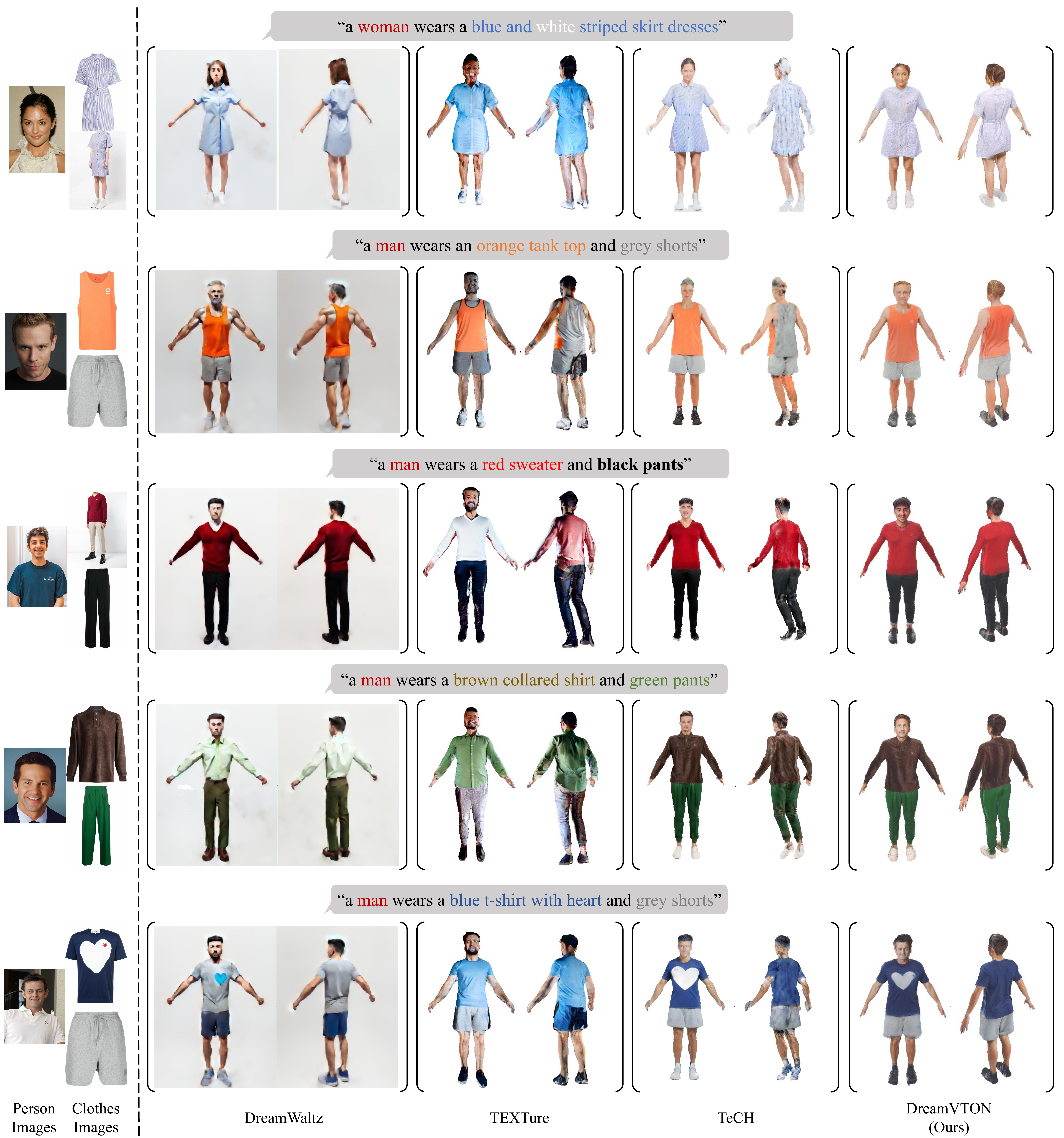}
  % \vspace{-6mm}
  \caption{Qualitative Comparisons. Using the same
clothes, person image, and prompt as inputs, our method achieves superior results.} 
  % \vspace{-6mm}
  \label{fig:vis_comp_v2}
\end{figure*}

We provide additional visual comparisons among our proposed DreamVTON and the existing 3D human generation methods (i.e., DreamWaltz~\cite{huang2023dreamwaltz}, TEXTure~\cite{richardson2023texture}, and TeCH~\cite{huang2024tech}) in Figure~\ref{fig:vis_comp_v1} and Figure~\ref{fig:vis_comp_v2}.

% \clearpage

%%
%% The next two lines define the bibliography style to be used, and
%% the bibliography file.
\bibliographystyle{ACM-Reference-Format}
\bibliography{dreamvton}

\end{document}